\documentclass[letterpaper, 10 pt, conference]{ieeeconf}  

\IEEEoverridecommandlockouts                              

\usepackage{color}
\usepackage{booktabs}
\usepackage{float}
\usepackage{graphicx}
\usepackage{array}
\usepackage{multirow}
\usepackage{makecell}
\usepackage{tabularx}
\usepackage{threeparttable}
\usepackage{caption}
\usepackage{subcaption}
\usepackage{cite}
\usepackage{xcolor}
\usepackage{url}

\definecolor{colorbox_yellow}{RGB}{255,255,0}
\definecolor{as_green}{RGB}{0,176,80}
\definecolor{ppt_orange}{RGB}{210,140,102}
\definecolor{fig1_purple}{RGB}{120,32,110}
\definecolor{fig1_dark_green}{RGB}{112,187,85}
\definecolor{fig1_light_blue}{RGB}{104,142,191}

\title{\LARGE \bf
\texttt{AutoSpatial}: Visual-Language Reasoning for Social Robot Navigation through Efficient Spatial Reasoning Learning
}

\author{Yangzhe Kong$^{1}$, Daeun Song$^{1}$, Jing Liang$^{2}$, Dinesh Manocha$^{2}$, Ziyu Yao$^{1}$, and Xuesu Xiao$^{1}$
\thanks{$^{1}$George Mason University 
        {\tt\small \{ ykong7, dsong26, ziyuyao, xiao \} @gmu.edu} $^{2}$University of Maryland, College Park 
        {\tt\small \{ jingl, dmanocha\}@umd.edu } }%
}

\begin{document}

\maketitle
\thispagestyle{empty}
\pagestyle{empty}

\begin{abstract}

We present a novel method, AutoSpatial, an efficient approach with structured spatial grounding to enhance VLMs’ spatial reasoning. By combining minimal manual supervision with large-scale Visual Question-Answering (VQA) pairs auto-labeling, our approach tackles the challenge of VLMs’ limited spatial understanding in social navigation tasks.
By applying a hierarchical two-round VQA strategy during training, AutoSpatial achieves both global and detailed understanding of scenarios, demonstrating more accurate spatial perception, movement prediction, Chain of Thought (CoT) reasoning, final action, and explanation compared to other SOTA approaches. These five components are essential for comprehensive social navigation reasoning. Our approach was evaluated using both expert systems (GPT-4o, Gemini 2.0 Flash, and Claude 3.5 Sonnet) that provided cross-validation scores and human evaluators who assigned relative rankings to compare model performances across four key aspects.
Augmented by the enhanced spatial reasoning capabilities, AutoSpatial demonstrates substantial improvements by averaged cross-validation score from expert systems in: perception \& prediction (up to 10.71\%), reasoning (up to 16.26\%), action (up to 20.50\%), and explanation (up to 18.73\%) compared to baseline models trained only on manually annotated data.


\end{abstract}

\section{INTRODUCTION}
\begin{figure}[!ht]
    \centering
    \includegraphics[width=\linewidth]{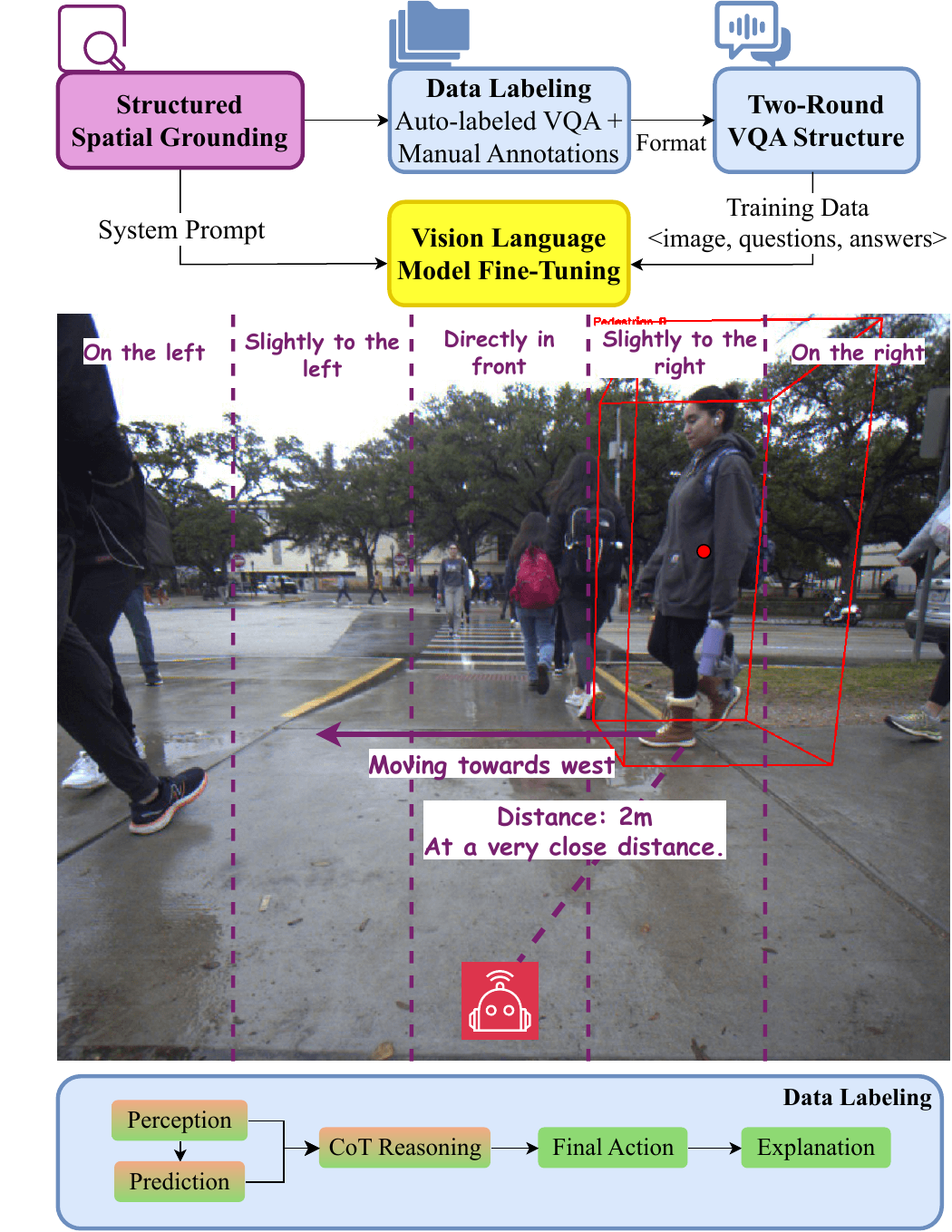}
    \caption{Overview of the AutoSpatial approach. The approach consists of two primary components: \textbf{{\color{fig1_purple} Structured Spatial Grounding}} (purple), which standardizes positional (e.g., `Slightly to the right' and `At a very close distance' for the pedestrian in the bounding box) and directional descriptions (e.g., `Moving towards west'), and \textbf{{\color{fig1_light_blue} Data Labeling}} (blue), which combines auto-labeled Visual Question-Answer (VQA) pairs that focus on {\color{ppt_orange}\bfseries percetion, prediction}, and {\color{ppt_orange}  \bfseries CoT reasoning}, with manual annotations that focus on {\color{fig1_dark_green} \bfseries percetion, prediction, CoT reasoning}, {\color{fig1_dark_green} \bfseries final action}, and  {\color{fig1_dark_green} \bfseries explanation}, and further formats them into a two-round VQA structure (see details in Fig.~\ref{fig:example_vqa}).}

    
    \label{fig:framework}
\end{figure}

Social navigation poses unique challenges for robots in human environments, where avoiding obstacles is insufficient—robots must also interpret and respond to nuanced human behaviors~\cite{mirsky2024conflict, mavrogiannis2023core}. While traditional and learning-based methods have improved navigation~\cite{helbing1995social, hirose2023sacson}, they often fail in dynamic social scenarios due to a lack of explicit spatial reasoning~\cite{epstein2017cognitive}.

Vision-Language Models (VLMs) offer promising reasoning capabilities~\cite{liu2024visual}, but their application to social navigation is hindered by two major limitations: (1) poor spatial grounding, as they lack explicit metrics like relative positions or movement trajectories~\cite{zhang2024vision, kamath2023s}; and (2) insufficient domain-specific data, since most annotated datasets focus on driving, not pedestrian interaction~\cite{sima2024drivelm, nejatishahidin2024structured}.

\textbf{Main Results: } To address these limitations, we introduce AutoSpatial, an efficient approach that systematically enhances spatial reasoning capabilities in VLMs for social navigation reasoning, including human's motion, relative position to the robot, and social interactions. AutoSpatial employs a structured question-answer paradigm to provide ground truth text labels for various visual observations and to train VLMs in perception, prediction, Chain-of-Thought (CoT) reasoning, final action, and explanation, enabling them to better interpret spatial relationships and social dynamics. Our approach leverages limited manual annotations in combination with large-scale auto-labeled data to significantly reduce annotation costs while maintaining high data quality.
We conduct extensive evaluations on the SNEI benchmark~\cite{payandeh2024social} for spatial reasoning and social action estimation
(Fig.~\ref{fig:example_error} right).
\begin{itemize}
    \item We propose a VLM-based approach, AutoSpatial, for enhancing spatial reasoning in social navigation;
    \item We introduce a novel automatic data labeling strategy with a two-round Visual Question-Answer (VQA) structure, inlcuding individual and group social motion understaning, that efficiently addresses domain-specific data scarcity while hierarchically enhancing spatial understanding;
    \item We evaluate our approach using averaged cross-validation scores from expert systems (GPT-4o, Gemini 2.0 Flash, Claude 3.5 Sonnet) and relative rankings from human evaluators. AutoSpatial outperforms other approaches  in perception \& prediction (up to 10.71\%), reasoning (up to 16.26\%), action (up to 20.50\%), and explanation (up to 18.73\%) in averaged cross validation of expert models.
\end{itemize}



\section{RELATED WORK}
VLMs, which combine visual perception with natural language understanding, have demonstrated significant potential across various robotics tasks. We review related work in social robot navigation and VLMs for robotics tasks, with particular focus on their spatial reasoning capabilities and limitations in dynamic social environments.

\subsection{Social Robot Navigation}

Robust social navigation is essential for robots in human-populated spaces~\cite{mirsky2024conflict, francis2023principles, mavrogiannis2023core}. Early approaches used hand-crafted rules, such as the Social Force Model~\cite{helbing1995social} and proxemics-based methods~\cite{daza2021approach}, which simulate human-robot interactions via spatial constraints. Learning-based methods~\cite{luber2012socially, hirose2023sacson, xiao2020appld, song2024vlm, raj2024rethinking} improve adaptability by leveraging large datasets to learn socially compliant behaviors. However, these models often lack explicit reasoning about nuanced human-robot dynamics shaped by unwritten social norms~\cite{payandeh2024social, chen2024spatialvlm}.

\subsection{Spatial Reasoning Limitations of Robotics VLMs}

Recent VLMs enable task and motion planning~\cite{brohan2023can, song2024tgs, nasiriany2024pivot} and support human-robot interaction~\cite{narasimhan2024olivia, song2024vlm}. Yet, two core challenges hinder their performance in social navigation:

\textbf{Lack of spatial reasoning:} VLMs lack explicit spatial encodings (e.g., trajectories, cost maps)~\cite{mavrogiannis2023core}, and struggle with ambiguous relative positioning~\cite{kamath2023s, wang2024picture, chen2024spatialvlm}.

\textbf{Insufficient domain-specific data:} Unlike other domains~\cite{cao2023instruction}, social navigation lacks large-scale annotated datasets reflecting complex, implicit crowd behavior~\cite{nejatishahidin2024structured, payandeh2024social, sima2024drivelm}.

We address these challenges with AutoSpatial, which augments VLMs through structured spatial VQA training, combining minimal manual annotation and large-scale auto-labeling.

\section{METHODOLOGY}
\begin{figure*}[!h]
        \centering
        \includegraphics[width=\textwidth]{images/2round_alt.pdf}
    \caption{An example of the two-round VQA structure, where training data of each round follows the same format of $<$image, question, ground-truth answer$>$. {\color{ppt_orange} \textbf{Round 1}} is auto-labeled, focusing on perception, prediction, and CoT reasoning for individual pedestrians, while {\color{fig1_dark_green} \textbf{Round 2}} is optional and only available when the image input is manually annotated. It refines individual descriptions into a comprehensive scene-level understanding, incorporating group dynamics and higher-level spatial reasoning. The structured spatial grounding, though not explicitly shown here, is incorporated into the system prompt as part of the model input. Note that bounding boxes and color-pedestrian correspondence are only provided to the model during training. 
    }
    \label{fig:example_vqa}
    \vspace{-10pt}
  \end{figure*}

We present AutoSpatial, a systematic approach for enhancing VLMs' spatial reasoning capabilities in social navigation. As illustrated in Fig.~\ref{fig:framework}, our approach is built upon the structured spatial grounding that standardizes positional and directional descriptions, providing a consistent spatial terminology foundation. Guided by this, we label our training data through two complementary approaches: auto-labeled VQA pairs for large-scale spatial understanding and targeted manual annotations for complex social scenarios. These data are further formatted into a structured two-round VQA architecture (see details in Fig.~\ref{fig:example_vqa}), which enables hierarchical learning from basic spatial perception to complex reasoning. This methodology addresses our key observation that existing VLMs often struggle with fundamental spatial reasoning, leading to incorrect actions even when high-level reasoning appears sound.

\subsection{Structured Spatial Grounding}  \label{sec:spatial_ref}


Our key innovation is to decompose spatial relationships into two clear, independent components: (1) positional description of humans, including angular position and distance, and (2) description of human-moving directions. This structured representation makes it easier to precisely locate and track pedestrians in relation to the robot. Traditional approaches, which describe spatial relationships between humans and robots in navigation, often lead to ambiguity, especially when addressing multiple agents \cite{zhang2024vision}.

For angular position of pedestrians, we divide the robot's field of view into five distinct zones: `on the left', `slightly to the left', `directly in front', `slightly to the right', and `on the right' of the robot. This discretization provides a clear way to describe a pedestrian's relative position w.r.t. the robot, while maintaining sufficient granularity for navigation tasks. Each pedestrian is assigned to a zone based on their center point's horizontal pixel location in the image.

Distance estimation follows a five-level classification: `very close' ($<3m$), `close' ($3-6m$), `moderate' ($6-9m$), `far' ($9-12m$), and `very far' ($>12m$). By combining these two components, our approach can generate precise spatial descriptions such as ``the pedestrian is slightly to the left of the robot at a moderate distance.''

For movement directions, we adopt a unified coordinate system relative to the robot's frame, where the robot's forward direction is defined as north. This allows us to describe pedestrian movements using eight cardinal and intercardinal directions (N, NE, E, SE, S, SW, W, NW) plus a stationary state. This standardization eliminates the ambiguity often found in relative directional descriptions and provides a consistent reference frame for perception and prediction.

The structured spatial grounding plays a critical role in both our auto-labeling process and manual annotation. During auto-labeling, it provides a consistent framework for generating spatial VQA pairs, ensuring that perception, prediction, and reasoning outputs are systematically structured. For manual annotations, it serves as a guideline for annotators to maintain consistency in descriptions, reducing subjective variations.

\subsection{Data Labeling} 

Our training data combines auto-labeled VQA pairs and manual annotations to achieve both breadth in spatial reasoning and depth in scene understanding. As shown in Fig.~\ref{fig:example_vqa}, the upper part illustrates the auto-labeled VQA pairs, while the lower part represents the manually annotatations. 
    \subsubsection{Auto-labeled VQA Pairs} We develop a systematic approach to generate VQA pairs based on the CODA dataset~\cite{zhang2023robust}. This dataset provides a rich collection of real-world pedestrian interactions in urban environments captured from a robot's perspective, offering the necessary trajectory information and bounding box annotations required by our approach. Note that this process relies solely on rule-based heuristics rather than complex learning-based models such as VLMs. For each frame, our approach processes three key aspects:

        \textbf{Spatial Perception}: Spatial perception consists of two components: positional description and movement direction description. Precise spatial descriptions are generated using our structured spatial grounding discussed in Sec.~\ref{sec:spatial_ref}.
        
        
        
        \textbf{Movement Prediction}: Our approach analyzes short-term trajectories using a multi-frame sliding window. After smoothing velocity vectors to reduce noise, it categorizes movement patterns into five primary types: started motion, continued motion, left turns, right turns, and transitions to stationary states. Predictions are generated based on velocity and direction changes exceeding predefined thresholds.
        
        \textbf{Interaction Description}: Our approach identifies, classifies, and describes various interaction patterns: 1. Trajectory conflicts: Detected by analyzing relative positions and velocities; 2. Path crossing: Categorized as west-to-east or east-to-west movements; 3. Pass-by scenarios: Classified based on relative trajectories; 4. Others: Categorized as scenarios without risks of collision.
    \subsubsection{Manual Annotations} 
    We carefully select 72 challenging scenarios from the CODA dataset, focusing on cases where auto-labeled VQAs show limitations. These scenarios typically involve high pedestrian density environments, complex social grouping patterns, and situations requiring abstraction from individual to group-level understanding, where auto-labeled VQAs tend to exhibit significant errors or ambiguities. 
    

    Following the SNEI protocol \cite{payandeh2024social}, we also annotate data using the structured format of 5 tasks: Perception, Prediction, CoT Reasoning, Final Action, and Explanation. However, our approach differs in that we provide our structured spatial grounding guideline for annotators, as described in~\ref{sec:spatial_ref}. This ensures stricter spatial terminology and reduced ambiguity, which not only improves annotation consistency but also provides clearer supervision.
\subsection{Two-Round VQA Structure} 
    
On top of the generated data, we use a two-round VQA structure (Fig.~\ref{fig:example_vqa}) for scenes with manual annotations to facilitate hierarchical learning. For each selected scene, the first round consists of auto-labeled VQAs focusing on individual pedestrians, while the second round incorporates human annotations to refine group dynamics, social relationships, and collective movement patterns. This hierarchical approach enhances spatial reasoning by first establishing precise spatial-temporal understanding through structured auto-labeling, followed by higher-level scene interpretation via human annotations. 

Specifically, we structure our data into consistent formats. For scenes with only auto-labeled data, we use a single-round format of $<$image, round 1 question, round 1 ground-truth answer$>$. For scenes with manual annotations, we structure the data as a two-round conversation: $<$image, round 1 question, round 1 ground-truth answer$>$ followed by $<$image, round 2 question, round 2 ground-truth answer$>$. 

This progressive structure yields synergistic effects beyond simple data augmentation. To explore this interplay, we conduct experiments (see Section \ref{sec:ablation}) to assess how spatial reasoning capability acquisition and social context comprehension mutually reinforce each other.

\section{EXPERIMENTS}
We conduct extensive experiments to evaluate the effectiveness of AutoSpatial in enhancing spatial reasoning for social navigation. Our evaluation focuses on assessing improvements in perception \& prediction, reasoning, action, and explanation, comparing our approach against baseline models.

\subsection{Experimental Setup}


We build our model based on LLaVA-1.6 OneVision-7B and conduct all experiments using a computing node with four A100 80GB GPUs. We follow the default instruction
tuning procedure and hyperparameters from LLaVA \cite{liu2024llavanext} without any special loss weighting between different data sources. Our focus is on evaluating the effectiveness of our data labeling and structured VQA framework rather than sophisticated training techniques. For each training sample, we provide the model with an input image along with corresponding VQAs following our structured spatial grounding guideline. We use the AdamW optimizer with a learning rate of $2e{-5}$ and batch size of $8$ with gradient accumulation steps of $4$. To comprehensively evaluate our approach, we establish a two-fold evaluation framework that assesses both basic spatial reasoning capabilities and high-level scene understanding.

For evaluating basic spatial reasoning, we employ an automated metric on the auto-labeled VQA pairs from CODA. Specifically, we first perform data balancing to mitigate directional bias by ensuring diverse pedestrian movement directions, then partition the dataset into training and test sets with a 90-10 split ratio.
This results in 6,205 training frames (containing 16,530 pedestrian instances) and 573 test frames (containing 1,542 pedestrian instances). The test set evaluation focuses on two fundamental aspects of spatial understanding that can be objectively assessed: movement direction prediction and relative position estimation. Given the inherent variability in text generation, we implement a keyword-based evaluation metric, where model outputs are matched against predefined categorical labels. The scoring system awards 1.0 point for exactly correct answers and 0.5 points for approximately correct answers (e.g., predicting ``northeast'' when the ground truth is ``north''). 

For comprehensive scene understanding evaluation, we use the SNEI benchmark with both expert systems, such as GPT-4o, Gemini 2.0 Flash, and Claude 3.5 Sonnet, and 7 human evaluators, most of whom are graduate computer science students with backgrounds in computer vision and robotics. Our human evaluation focused on a randomly sampled subset of 49 scenarios from the SNEI dataset. Both expert systems and human evaluators assess four key aspects---Perception \& Prediction, Reasoning, Action, and Explanation. Expert systems assign scores on a scale of 1-10, while human evaluators focus on our best method alongside two baseline models and provide relative rankings for each key aspect. We chose relative rankings for human evaluation rather than absolute scores because we believe it can mitigate potential subjective biases. For a fair comparison, we ensure all models receive identical input formats and evaluation criteria.


\begin{table*}[ht]
\centering
\vspace{1em}
\begin{threeparttable}
\caption{Cross-Validation Scores across Models for Different Aspects (Higher is Better).}
\renewcommand{\arraystretch}{1.3}
\begin{tabular}{c|c|cccc|cccc|cccc}
\toprule
\multirow{2}{*}{\textbf{Model ID}\tnote{1}} & \multirow{2}{*}{\makecell[c]{\textbf{CODA}\\  \textbf{Benchmark}}} 
& \multicolumn{4}{c|}{\textbf{Gemini 2.0 flash}} 
& \multicolumn{4}{c|}{\textbf{GPT-4o}} 
& \multicolumn{4}{c}{\textbf{Claude 3.5 Sonnet}} \\
\cline{3-14}
& & P\&P & R & A & E & P\&P & R & A & E & P\&P & R & A & E \\
\midrule
\textbf{AS-A72D1} &\textbf{0.710}&6.824&\textbf{7.079}&7.292&7.258 & 6.838&\textbf{6.338}&6.290&\textbf{6.471} & 6.652&\textbf{5.831}&\textbf{6.385}&\textbf{6.065} \\
\textbf{AS-A72D2} & 0.582&\textbf{6.883}&7.070&7.272&7.264 & \textbf{6.862}&6.221&\textbf{6.302}&6.413 & \textbf{6.677}&5.671&6.351&6.000 \\
\textbf{AS-A72D3} & 0.573&6.775&7.048&\textbf{7.360}&\textbf{7.310} & 6.813&6.077&6.250&6.269 & 6.674&5.591&6.292&5.957 \\
\textbf{AS-A72D5} & 0.483&6.627&6.878&7.280&7.270 & 6.739&5.947&6.114&6.139 & 6.504&5.490&6.151&5.831 \\
\textbf{AS-A36D1} & 0.686&6.661&6.996&7.277&7.227 & 6.764&6.135&6.167&6.247 & 6.584&5.613&6.234&5.936 \\
\textbf{AS-A0D1} &
0.680&6.484&6.660&6.859&6.756 & 6.582&5.517&5.733&5.920 & 6.551&5.415&4.831&5.375 \\
\midrule
\textbf{LLaVA-M} & 0.404&6.246&6.089&6.108&6.157 & 6.662&5.815&5.791&6.351 & 6.031&5.065&5.342&5.302 \\
\textbf{Vanilla-LLaVA} & 0.376&6.243&6.083&6.102&6.154 & 6.486&5.492&5.458&6.092 & 5.809&4.815&4.803&5.409 \\
\bottomrule
\end{tabular}
\label{tab:cross_agreement}
\begin{tablenotes}
    \footnotesize \item[1] The model IDs follow the format `AS-AxxDy', where `AS' denotes AutoSpatial, `Axx' indicates the number of manual annotations used (e.g., A72 for 72 annotations), and `Dy' represents the downsampling ratio
applied to the auto-labeled data (e.g., D1 for no additional downsampling).
\end{tablenotes}
\end{threeparttable}
\end{table*}

\begin{table*}[!ht]
\centering
\caption{Ablation Study}
\renewcommand{\arraystretch}{1.3}
\begin{tabular}{c|c|cccc|cccc|cccc}
\toprule
\multirow{2}{*}{\textbf{Model ID}} & \multirow{2}{*}{\makecell[c]{\textbf{CODA}\\  \textbf{Benchmark}}} 
& \multicolumn{4}{c|}{\textbf{Gemini 2.0 flash}} 
& \multicolumn{4}{c|}{\textbf{GPT-4o}} 
& \multicolumn{4}{c}{\textbf{Claude 3.5 Sonnet}} \\
\cline{3-14}
& & P\&P & R & A & E & P\&P & R & A & E & P\&P & R & A & E \\
\midrule
\textbf{AS-A72D1-P\&P-VQA} &0.709&\textbf{6.781}&7.014&7.203&7.153 & 6.832&6.194&6.185&6.004 & \textbf{6.652}&5.717&6.375&6.010 \\
\textbf{AS-A72D1-R-VQA-1} & 0.518&6.717&7.008&7.246&7.206 & 6.601&5.541&5.767&5.542 & 6.621&5.684&6.323&5.982 \\
\textbf{AS-A72D1-R-VQA-2} & 0.520&6.751&\textbf{7.126}&\textbf{7.287}&\textbf{7.246} & \textbf{6.835}&\textbf{6.231}&\textbf{6.287}&\textbf{6.010} & 6.584&\textbf{5.853}&\textbf{6.397}&\textbf{6.050} \\
\midrule
\textbf{AS-A72D1}&0.710&6.824&7.079&7.292&7.258 & 6.838&6.338&6.290&6.065 & 6.652&5.831&6.385&6.065 \\
\bottomrule
\end{tabular}
\label{tab:ablation_study}
\end{table*}

\begin{table}[ht]
\centering
\caption{Human Evaluator Rankings for Different Models.}
\renewcommand{\arraystretch}{1.3}
\begin{tabular}{c|cccc}
\toprule
\textbf{Model} & \textbf{P\&P} & \textbf{R} & \textbf{A} & \textbf{E} \\
\midrule
\textbf{AS-A72D1}       & \textbf{1.19}  & \textbf{1.70}  & \textbf{1.43}  & \textbf{1.55}  \\
\textbf{LLaVA-M}   & 2.89  & 2.02  & 2.03  & 1.73  \\
\textbf{Vanilla LLaVA}  & 2.93  & 2.03  & 2.55  & 2.53  \\
\bottomrule
\end{tabular}
\label{tab:human_eval}
\vspace{-10pt}
\end{table}

To analyze different aspects of our approach, we compare several model configurations. Our baseline models include the original LLaVA-1.6 OneVision-7B and LLaVA finetuned on only 72 manual annotations(referred to as LLaVA-M, where 'M' denotes 'Manual annotations'). We deliberately focus on comparing with these 7B-parameter models to ensure a fair comparison, as larger proprietary models with hundreds of billions of parameters naturally demonstrate superior performance but operate at a fundamentally different scale, which prohibits inference onboard a mobile robot. We then evaluate variations of our approach with different training data compositions: models trained, with the full dataset combining 72 manual annotations and CODA VQA data (18,615 pairs in total), and with different overall downsampling ratios (AS-A72D1 as 1, AS-A72D2 as 2, AS-A72D3 as 3, and AS-A72D5 as 5, see Tab.~\ref{tab:cross_agreement}) of CODA VQA data, to examine the impact of different data mixing ratios. To investigate the impact of manual annotation quantity, we also train models with reduced manual annotations (36 examples, see AS-A36D1 in Tab.~\ref{tab:cross_agreement}) and with no manual annotations (AS-A0D1). Additionally, we conduct ablation studies using only perception \& prediction VQAs (AS-A72D1-P\&P-VQA) and only reasoning VQAs with manual annotations (AS-A72D1-R-VQA-1 and AS-A72D1-R-VQA-2) to understand their individual contributions.

\begin{figure*}[!h]
\vspace{0.5em}
    \centering
    \begin{subfigure}{0.2\textwidth}
        \centering
        \includegraphics[width=\textwidth]{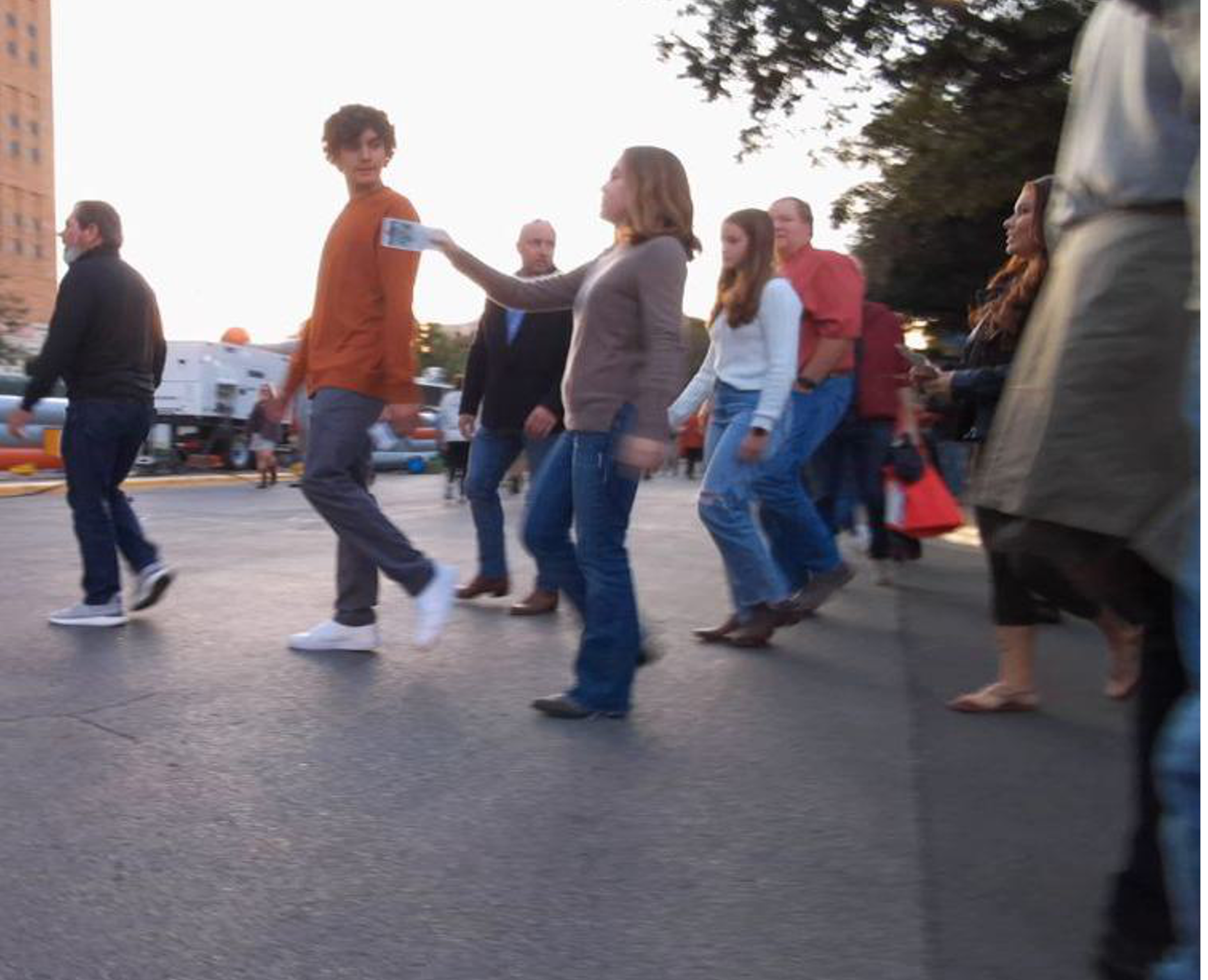}
        \vspace{20mm}
        \caption{Input Image}
    \end{subfigure}
    \begin{subfigure}{0.79\textwidth}
        \centering
        \includegraphics[width=\textwidth]{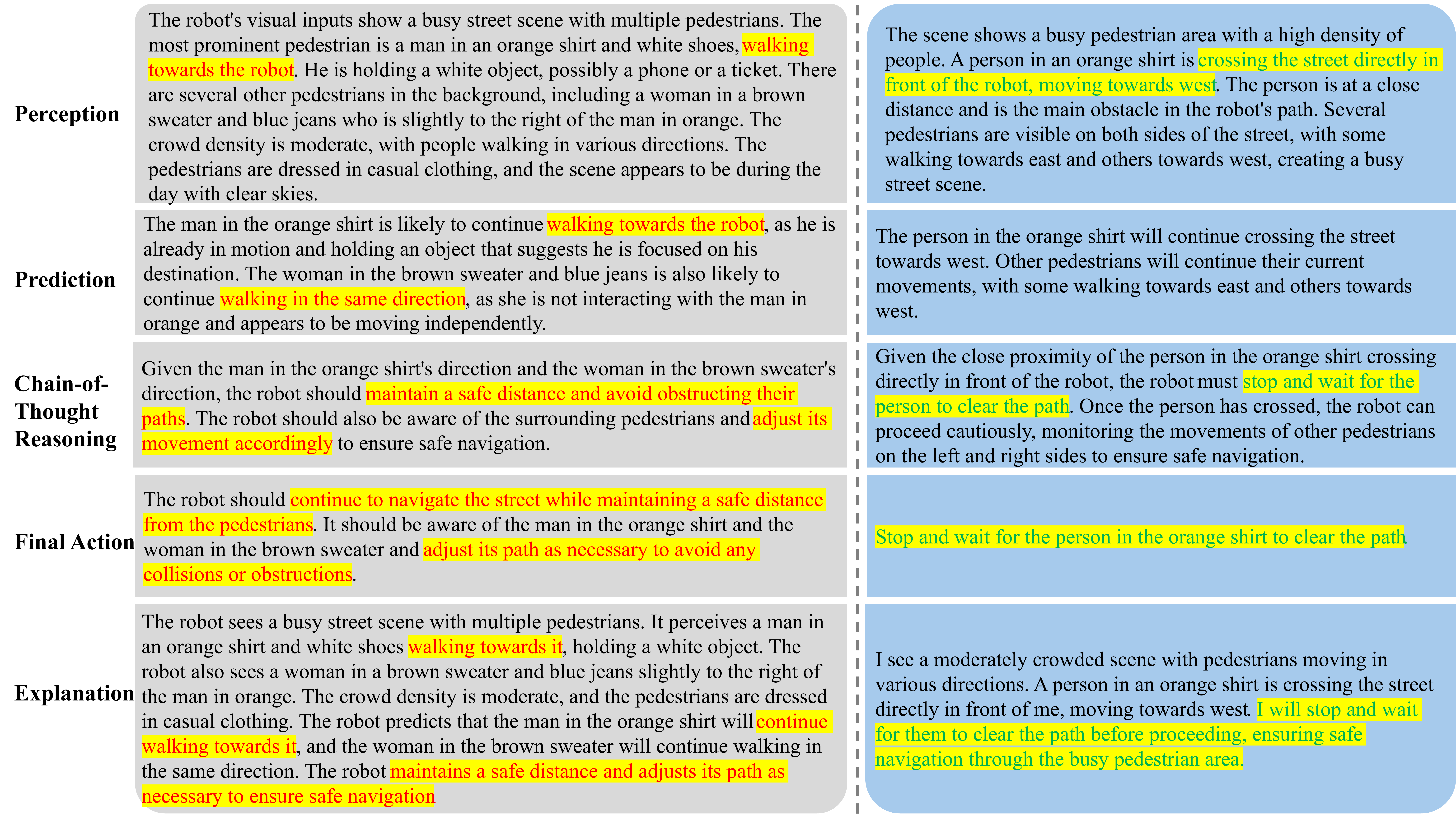}
        \caption{Output comparison: LLaVA-M (left) v.s. AutoSpatial (right)}
    \end{subfigure}
    \caption{
    %
    While LLaVA-M suffers from {\setlength{\fboxsep}{1pt}\colorbox{yellow}{{\color{red}{faulty spatial reasoning, leading to ambiguous }}}} {\setlength{\fboxsep}{1pt}\colorbox{yellow}{{\color{red}{or ineffective navigation decisions}}}}, AutoSpatial exhibits {\setlength{\fboxsep}{1pt}\colorbox{yellow}{{\color{as_green}improved pedestrian identification and reasoning}}}, when augmented with auto-labeled VQA pairs.
    } 
    \label{fig:example_error}
\end{figure*}

\subsection{Results}


Tab.~\ref{tab:cross_agreement} and Tab.~\ref{tab:human_eval} present a comprehensive evaluation of our approach through both automated metrics and human assessment. We evaluate all model variants on the CODA benchmark for basic spatial reasoning and using three state-of-the-art VLMs (Gemini 2.0 flash, GPT-4o, and Claude 3.5 Sonnet) for expert assessment on SNEI dataset, focusing on four aspects: Perception \& Prediction (P\&P), Reasoning (R), Action (A), and Explanation (E). Additionally, we conduct focused human evaluation on the three most representative models: our best-performing variant (AS-A72D1, AutoSpatial with 72 human Annotations and Downsampling ratio 1), LLaVA-M, and vanilla LLaVA.

Our proposed method (AS-A72D1) demonstrates consistent superior performance across all evaluation metrics. In the CODA benchmark focusing on spatial reasoning, AS-A72D1 achieves a score of 0.710, significantly outperforming both LLaVA-M (0.404) and vanilla LLaVA (0.376). This substantial improvement in spatial reasoning capabilities is further validated by both expert system evaluations and human assessment (Tab.~\ref{tab:human_eval}), where AS-A72D1 consistently outperforms the baseline models, particularly in perception \& prediction and reasoning capabilities. 


A key finding is that manual annotations and auto-labeled data complement each other, and combining them yields a synergetic effect, surpassing the performance of models trained on either data source alone. While LLaVA-M outperforms Vanilla-LLaVA, the gap remains substantial, indicating that manual annotations alone are insufficient to fully address the spatial reasoning deficiencies in VLMs. Notably, training on only auto-labeled VQA data (AS-A0D1) achieves only a marginal performance improvement over Vanilla-LLaVA and, in some cases, even underperforms, demonstrating that human supervision is still essential for grounding social interactions. 


The robustness of our approach is demonstrated through several key findings. First, varying the downsampling ratio (AS-A72D1 through AS-A72D5) results in only minor performance differences, indicating that our method is relatively insensitive to the exact mixing proportions of the training data. Second, reducing manual annotations by half (AS-A36D1) leads to minimal performance degradation, suggesting that our approach can achieve strong results even with limited manual supervision.


\subsection{Ablation Studies} \label{sec:ablation}

Tab.~\ref{tab:ablation_study} examines the individual contributions of different components in our framework. Training with only perception and prediction VQA tasks (AS-A72D1-P\&P-VQA) shows strong performance in the P\&P metric but limited improvement in other aspects. Conversely, using only reasoning VQA tasks without the ground truth of perception and prediction VQA tasks (AS-A72D1-R-VQA-1) improves reasoning, action, and explanation scores but struggles with basic spatial understanding. Even with the ground truth of the perception and prediction VQA tasks provided in the question (AS-A72D1-R-VQA-2), the improvement in P\&P remains limited.

These ablation results reveal a crucial insight: the synergistic effect of combining different types of VQAs with manual annotations. While individual VQA types can improve specific aspects of performance, only the full combination achieves optimal results across all metrics. 



\subsection{Qualitative Analysis}
To better understand our model's capabilities and limitations, we analyze one representative case that highlights the differences between our approach and baseline models. In the scenario shown in Fig.~\ref{fig:example_error}, featuring a crowded sidewalk with multiple pedestrians, our model demonstrates superior spatial awareness by clearly identifying key actors and their relative positions using standardized spatial terminology (e.g., ``moving towards west'' instead of ``walking towards/away from the camera''). More importantly, while baseline models generate generic advice like ``continue to navigate the street while maintaining a safe distance'', our model provides specific, socially-aware instructions---recommending to wait for pedestrians to cross before proceeding, which better aligns with social norms.

However, our analysis also reveals areas for improvement. While the model demonstrates strong spatial reasoning, it occasionally misinterprets subtle human cues, such as gaze direction and body orientation, which are crucial for understanding pedestrian intent. Additionally, it sometimes reverts to ambiguous camera-relative descriptions used intensively by LLaVA instead of consistently applying the structured spatial grounding, likely due to data limitations during training.



This qualitative analysis validates our approach's effectiveness in improving spatial reasoning while also identifying clear directions for future enhancement. The improved spatial reasoning capabilities directly contribute to more specific and socially appropriate navigation decisions, even though opportunities remain for further refinement in human behavior understanding.

\section{CONCLUSIONS}

In this paper, we present an efficient approach to enhance VLMs' spatial reasoning capabilities through a combination of auto-labeled data and minimal manual annotations. Our method demonstrates that with just 72 carefully selected manual annotations combined with structured auto-labeled data, we can achieve significant improvements in both basic spatial understanding and high-level scene comprehension.

Our experimental results reveal several key findings. First, the combination of auto-labeled VQAs focusing on individual pedestrians and manual annotations emphasizing group dynamics creates a synergistic effect that exceeds the performance of either approach alone. Second, our method shows remarkable robustness to various training configurations, maintaining strong performance even with reduced manual annotations. Third, qualitative analysis demonstrates that enhanced spatial reasoning capabilities directly contribute to more specific and socially appropriate navigation decisions.

\section*{ACKNOWLEDGMENT}
This work has taken place in the RobotiXX Laboratory at George Mason University. RobotiXX research is supported by National Science Foundation (NSF, 2350352), Army Research Office (ARO, W911NF2320004, W911NF2420027, W911NF2520011), Air Force Research Laboratory (AFRL), US Air Forces Central (AFCENT), Google DeepMind (GDM), Clearpath Robotics, Raytheon Technologies (RTX), Tangenta, Mason Innovation Exchange (MIX), and Walmart.

\bibliographystyle{IEEEtran}
\bibliography{IEEEabrv,ref.bib}

\end{document}